\documentclass[lettersize,journal]{IEEEtran}
\usepackage{amsmath,amsfonts}
\usepackage{booktabs}
\usepackage{multirow}
\usepackage{graphicx}
\usepackage{algorithmic}
\usepackage{algorithm}
\usepackage{placeins}
\usepackage{makecell}
\usepackage{cuted}
\usepackage{capt-of}
\usepackage{array}
\usepackage[caption=false,font=normalsize,labelfont=sf,textfont=sf]{subfig}
\usepackage{textcomp}
\usepackage{stfloats}
\usepackage{url}
\usepackage{verbatim}
\usepackage{graphicx}
\usepackage{cite}
\begin{document}

\title{XAI-Guided Fourier Domain Generalisation for Pure-Simulation Sim-to-Real Deep Reinforcement Learning}

\author{
Wei~Yao, Chuyao~Wang$^{*}$, Ziwei~Wang, and Nabil~Aouf\\[0.3em]
{\normalfont\normalsize City St George's, University of London}\\
{\normalfont\normalsize London, United Kingdom}\\[0.3em]
{\normalfont\small
Email: wei.yao, chuyao.wang.2, ziwei.wang.3, nabil.aouf@city.ac.uk,}

\thanks{*Corresponding author}
}

\markboth{Journal of \LaTeX\ Class Files,~Vol.~14, No.~8, 2026}%
{Shell \MakeLowercase{\textit{et al.}}: A Sample Article Using IEEEtran.cls for IEEE Journals}


\maketitle

\begin{abstract}
Deep reinforcement learning has shown strong potential for vision-based autonomous driving, but policies trained in simulation often suffer severe performance degradation when deployed in the real world. Existing sim-to-real methods commonly address this gap through image-transfer-based domain adaptation or policy-agnostic domain randomisation. However, the former requires laborious acquisition of target-domain data and is often limited to the deployment domain, whereas the latter introduces unguided perturbations that may not reflect the underlying sim-to-real discrepancy, leading to suboptimal transfer performance. To overcome these limitations, we propose XFD, an explainable artificial intelligence (XAI)-guided Fourier domain generalisation framework for pure-simulation sim-to-real reinforcement learning. The proposed framework extracts explainability attribution maps from simulated observations, and converts these attributions into policy-conditioned Fourier-domain interventions. By perturbing appearance-sensitive spectral components while preserving task-defining spatial structures, XFD constructs robustness-enhancing visual domains entirely within simulation. Subsequently, the proposed robust policy optimisation guides the policy towards attending to task-critical features while suppressing sensitivity to irrelevant visual information. Extensive experiments in simulation and zero-shot real-world deployment demonstrate that XFD consistently improves traveled distance and reward over standard policy and randomly domain-randomised baseline, especially on unseen maps with cluttered backgrounds. Attribution-based analysis further proves that the proposed policy preserves more consistent task-relevant visual evidence across simulation and real-world observations.
\end{abstract}

\begin{IEEEkeywords}
Autonomous driving, deep reinforcement learning, explainable artificial intelligence, sim-to-real transfer, visual domain generalisation.
\end{IEEEkeywords}

\section{Introduction}
\IEEEPARstart{A}{utonomous} driving is a closed-loop decision-making problem in which perception and control must remain reliable under uncertain and changing environments. Visual scene understanding and deep reinforcement learning have shown strong potential for autonomous driving, particularly when high-dimensional visual observations must be interpreted and mapped to low-level control actions\cite{0125c4a331104e63939b1737791ba1a2,10.1109/TITS.2020.3024655,9346000,9827377}. However, training such policies directly on physical robots\cite{8957584} is often unsafe, sample-inefficient, and difficult to reproduce. Simulation therefore provides a scalable and controllable platform for policy learning. Nevertheless, policies optimised in simulation often suffer performance degradation after deployment, because real-world visual appearance, camera response, vehicle dynamics, and environmental statistics may differ from those observed during training\cite{11181076}. This sim-to-real gap remains a central obstacle to robust learning-based autonomous navigation~\cite{gulrajani2021in,haarnoja2018softactorcriticoffpolicymaximum,tobin2017domainrandomizationtransferringdeep}.

Sim-to-real transfer requires sufficient resemblance between the training and deployment environments in visual appearance, scene statistics, sensor response, and robot dynamics~\cite{JRC128839}. Yet constructing a target-faithful simulator is rarely feasible, since real deployment involves uncontrolled illumination, texture variation, sensing noise, and hardware-dependent dynamics~\cite{DBLP:journals/corr/abs-2111-00956}. Rather than requiring an exact reconstruction of the target environment, domain randomisation relaxes this requirement by training the policy over a distribution of randomised simulation domains. Existing randomisation strategies can be broadly categorised into target-informed and target-free settings. Target-informed methods exploit target-domain evidence, including real images~\cite{DBLP:journals/corr/abs-1809-06256}, real-world rollouts~\cite{8793789}, or domain adaptation~\cite{bousmalis2017usingsimulationdomainadaptation}, to calibrate the randomised distribution toward the deployment domain. Target-free methods instead assume no access to target-domain data and enlarge visual, physical, or dynamical variability in simulation, aiming to include the real domain within the induced training support~\cite{tobin2017domainrandomizationtransferringdeep,doi:10.1177/0278364919887447}. Both settings seek to improve the overlap between training and deployment distributions, but differ in whether this overlap is guided by target-domain evidence or approximated through source-domain diversity.

The above resemblance-driven and randomisation-based strategies are commonly implemented at the image or pixel level, either by aligning visual appearance with the target domain or by perturbing simulator-induced visual factors. Although such pixel-level interventions can reduce apparent domain discrepancy, they entangle style-dependent appearance with task-defining spatial structure and provide limited control over which visual changes affect closed-loop behavior. Frequency-domain adaptation provides a more structured way to control visual shifts. Spectral amplitude is closely related to style-dependent factors such as texture, illumination, and color statistics, whereas phase information better preserves geometry and spatial structure~\cite{yang2020fdafourierdomainadaptation,xu2021fourierbasedframeworkdomaingeneralization}. However, existing Fourier-based methods usually borrow target-domain spectra or mix spectra across observed domains~\cite{9489280}, and the generated shifts are not explicitly linked to closed-loop policy behavior. To make spectral interventions policy-sensitive, it is necessary to identify which visual evidence contributes to the action decision. Attribution methods such as Grad-CAM and DeepLIFT provide such a mechanism by revealing the visual evidence behind model outputs~\cite{Selvaraju_2019,shrikumar2019learningimportantfeaturespropagating}. Nevertheless, these explanations are still mainly used for post-hoc analysis~\cite{10475532}, and explanation-guided learning has rarely been exploited to construct simulation interventions for sim-to-real reinforcement learning~\cite{gao2024egl}. As a result, existing methods remain limited in connecting structured visual shifts with the policy-relevant evidence that determines closed-loop behavior.

This paper proposes XFD, an XAI-guided Fourier domain generalisation framework for target-free sim-to-real reinforcement learning. The framework replaces policy-agnostic randomisation with policy-sensitive domain intervention: a frozen teacher policy first provides attribution maps that identify action-relevant visual evidence, and these maps are then used to modulate Fourier-domain perturbations generated entirely within simulation. Since training under target-free visual domain shifts can destabilise policy optimisation, the teacher is also used as an auxiliary stabilising reference~\cite{10670293}, rather than as an imitation target or performance upper bound. The student policy remains optimised by the reinforcement learning objective under these policy-conditioned domains, enabling robust policy improvement while preserving task behavior. Soft Actor-Critic (SAC)~\cite{haarnoja2018softactorcriticoffpolicymaximum} is adopted as the baseline reinforcement learning algorithm.

The main contributions of this work are summarised as follows:
\begin{itemize}
\item We propose a DeepLIFT-based XAI-guided Fourier domain intervention mechanism for visual generalisation. The proposed approach translates policy attributions into structured spectral perturbations that modify deployment-sensitive appearance statistics while preserving task-defining spatial structures, enabling robustness-enhancing visual domain generalisation entirely within simulation.
\item We introduce a target-domain-free robust policy optimisation strategy within a teacher-student framework. The teacher provides attribution guidance and training stabilisation rather than imitation targets, enabling the student policy to surpass the teacher in task performance while learning more focused and task-relevant representations across visual domains.
\item We conduct extensive evaluations in both simulation and zero-shot real-world deployment using the Duckietown platform. Results on lane-following and obstacle-avoidance tasks demonstrate that XFD consistently improves sim-to-real transfer performance on unseen maps with substantial background clutter, achieving robust generalisation through pure simulation training without requiring target-domain data acquisition or real-world fine-tuning.
\end{itemize}

\section{Related work}

\subsection{Sim-to-Real Visual Domain Generalisation}

Sim-to-real reinforcement learning has been widely studied to reduce the cost and risk of training visual control policies directly on physical robots. A common line of work narrows the sim-to-real gap by incorporating target-domain information. Bousmalis et al.~\cite{bousmalis2017usingsimulationdomainadaptation} combined randomised simulation with domain adaptation for robotic grasping from monocular RGB images, while Chebotar et al.~\cite{8793789} proposed SimOpt to adapt simulator parameter distributions using a small number of real-world rollouts. These methods improve transfer by aligning simulation with the deployment domain, but they still rely on real observations, real interactions, or target-domain statistics. Such dependence increases data-collection cost and may restrict transferability to the deployment conditions covered by the collected target data.

Visual appearance transfer addresses sim-to-real transfer from the perspective of image-domain alignment. Pixel-domain methods directly transform synthetic images toward a target visual style. Shrivastava et al.~\cite{shrivastava2017learningsimulatedunsupervisedimages} proposed SimGAN to refine simulator outputs using adversarial learning, Bousmalis et al.~\cite{bousmalis2017unsupervisedpixelleveldomainadaptation} studied unsupervised pixel-level domain adaptation, and Hoffman et al.~\cite{hoffman2017cycadacycleconsistentadversarialdomain} introduced CyCADA by combining image-level and feature-level adaptation with semantic consistency. Similar photorealistic translation has also been explored for synthetic driving scenes, such as CARLA2Real~\cite{Pasios_2025}. These methods reduce visual discrepancy between synthetic and real data, but they usually assume access to target-domain images and optimise image realism, feature alignment, or perception accuracy rather than closed-loop policy robustness. More importantly, unconstrained pixel-level translation may alter action-critical structures such as lane geometry, road boundaries, and obstacle layout.

Fourier-domain methods provide a more structured alternative by manipulating image spectra rather than pixels directly. Yang and Soatto~\cite{yang2020fdafourierdomainadaptation} proposed Fourier Domain Adaptation, which replaces the low-frequency amplitude spectrum of source images with that of target images. Xu et al.~\cite{xu2021fourierbasedframeworkdomaingeneralization} further showed that phase information tends to preserve semantic and geometric structure, whereas amplitude statistics are more closely related to domain-dependent appearance such as color, texture, and illumination. Compared with pixel-domain translation, Fourier-domain manipulation offers a clearer separation between appearance variation and structural preservation. However, existing Fourier-based methods either rely on target-domain spectra or perform generic spectrum mixing, and their perturbations remain disconnected from the visual evidence used by a closed-loop policy. This motivates a policy-aware use of Fourier-domain generalisation, where spectral-domain interventions are regulated by action-relevant visual evidence to modify deployment-sensitive appearance statistics while preserving task-defining spatial structure.

\subsection{Policy-Relevant Domain Construction}

Domain construction concerns how to generate source-domain variations that are not only diverse, but also effective for improving transfer robustness. Early domain randomisation methods broaden the simulation distribution by manually varying visual or physical factors. Tobin et al.~\cite{tobin2017domainrandomizationtransferringdeep} introduced visual domain randomisation by training models with randomised textures, distractors, camera parameters, lighting conditions, and image noise. Peng et al.~\cite{Peng_2018} extended this idea to control by randomising dynamics parameters during simulation training. These methods established the effectiveness of distributional broadening, but the induced domains are largely determined by predefined simulator parameters and sampling ranges.

Recent studies have moved toward objective-guided domain construction, where the randomised distribution is adapted according to a task- or transfer-related criterion. Bayesian domain randomisation searches source-domain distribution parameters to improve real-world performance using sparse target-domain data\cite{9327467}. Active domain randomisation selects informative environment instances according to trajectory-level discrepancy or task difficulty\cite{DBLP:journals/corr/abs-1904-04762}, while automatic and entropy-based randomisation progressively expands the domain distribution according to policy performance or success constraints~\cite{openai2019solvingrubikscuberobot,tiboni2024domain}. These methods show that domain generation should be guided by learning objectives rather than by uniform sampling alone. However, their objectives are typically defined over simulator parameters, distributional spread, trajectory-level discrepancy, real-world rollouts, or episode-level outcomes.

For visual closed-loop policies, such objective signals remain indirect: they indicate whether a domain is useful at the task level, but do not reveal which visual evidence drives the policy's action selection. As a result, existing objective-guided randomisation may improve domain coverage or task-level robustness without explicitly controlling whether the generated visual shifts affect policy-relevant perception. This motivates a policy-relevant domain construction strategy, where action-related visual evidence is used to guide robustness-enhancing visual shifts generated entirely within simulation.

\subsection{Explanation-Guided Training}

Attribution-based explanation methods have been widely used to analyse the visual evidence behind model predictions and XAI has become an important direction for safe and trustworthy autonomous driving\cite{10716567}. Model-agnostic methods such as LIME~\cite{DBLP:journals/corr/RibeiroSG16} approximate local decision behavior by perturbing input samples, while SHAP and its deep variants, such as DeepSHAP, estimate feature contributions from the perspective of Shapley values~\cite{DBLP:journals/corr/LundbergL17}. Gradient-based methods provide a more direct way to attribute model outputs to input features. Sundararajan et al.~\cite{DBLP:journals/corr/SundararajanTY17} proposed Integrated Gradients, which accumulates gradients along a path from a reference input to the actual input. Selvaraju et al.~\cite{Selvaraju_2019} proposed Grad-CAM, which produces class-discriminative localisation maps from convolutional feature maps. Shrikumar et al.~\cite{shrikumar2019learningimportantfeaturespropagating} proposed DeepLIFT, which propagates contribution scores relative to a reference activation and assigns dense input-level relevance to model outputs.

These methods differ in their suitability for closed-loop visual reinforcement learning. Perturbation-based and Shapley-value methods are general but often require repeated input evaluations or background sampling, making them less suitable for frequent attribution extraction during policy training. Grad-CAM is efficient and intuitive, but its localisation is usually coarse because it is derived from high-level feature maps. Integrated Gradients provides input-level attribution, but its path integration can be computationally more expensive. In contrast, DeepLIFT provides dense input-level contributions through reference-based propagation and is less affected by gradient saturation. This makes it suitable for identifying fine-grained visual evidence that influences continuous action outputs.

Beyond post-hoc interpretation, explanation-guided learning uses explanations as regularisation, supervision, or intervention signals to influence model training~\cite{gao2024egl}. In reinforcement learning, Zuo et al.~\cite{zuo2025agentdecisioncontrastiveexplanation} studied contrastive explanation learning for explaining agent decisions. Existing works show that explanations can reveal decision-relevant evidence in visual policies, but they are mainly used for interpretation, debugging, or explanation-level supervision. In contrast, this work uses DeepLIFT-based attribution as a training-time signal for domain construction, linking action-relevant visual evidence with policy-conditioned visual domain shifts for target-free sim-to-real transfer.

\section{Methodology}

\subsection{Overview}

\begin{figure*}[!t]
\centering
\includegraphics[
    width=0.95\textwidth,
    trim=0.2cm 3cm 0cm 0cm,
    clip
]{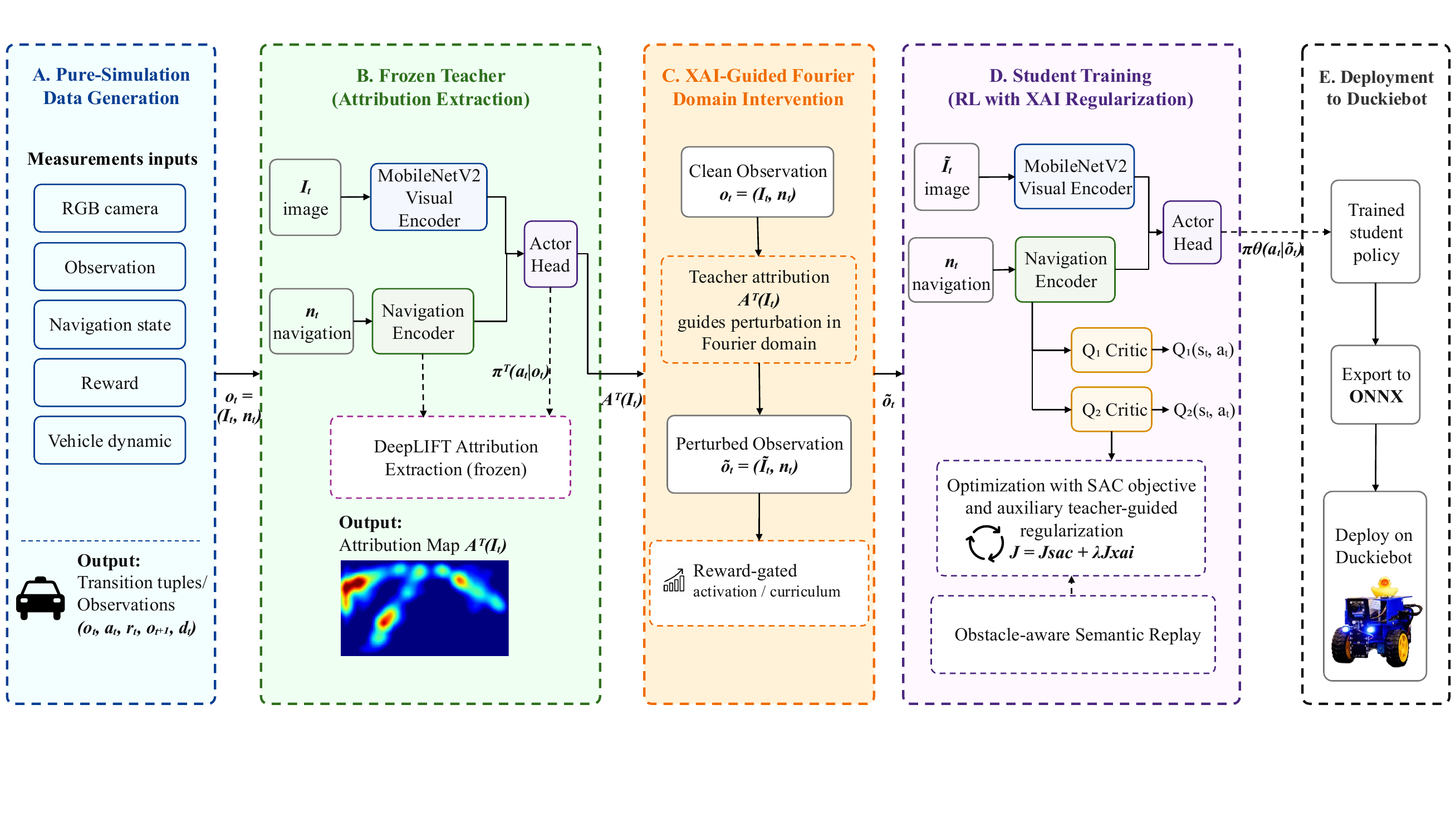}
\caption{Overall framework of the proposed target-free XAI-guided Fourier domain generalisation method. A single obstacle-aware policy is trained in simulation with randomised obstacle placements, where teacher attributions guide Fourier-domain visual intervention and semantic replay improves sampling of sparse obstacle-interaction transitions. The trained student policy is exported to ONNX for zero-shot Duckiebot deployment.}
\label{fig:framework}
\end{figure*}

The proposed framework aims to construct policy-relevant visual domain shifts entirely within simulation for target-free sim-to-real reinforcement learning. As illustrated in Fig.~\ref{fig:framework}, the framework follows a teacher-student structure. The teacher is a trained domain-randomised SAC policy and is kept frozen throughout training. The student is trained from sratch. During training, the student receives the XAI-guided Fourier-perturbed observations and is optimised by the proposed target-free robust policy objective. This student policy is indicated as the  XFD model used for simulation evaluation and zero-shot Duckiebot deployment. 

Based on the teacher attribution, the input image for the student model is perturbed in the Fourier domain. Following Fourier-domain adaptation \cite{yang2020fdafourierdomainadaptation} and Fourier-based domain generalisation\cite{xu2021fourierbasedframeworkdomaingeneralization}, the amplitude component is modified to introduce appearance variations related to illumination, texture, contrast, and color statistics, while the phase component is preserved to maintain lane geometry, road boundaries, and obstacle layout. The attribution map regulates the spatial expression of the perturbation, so that task-critical regions are protected and less relevant regions can absorb stronger appearance shifts. This converts generic spectral augmentation into a policy-conditioned domain intervention.

To maintain stable policy optimisation, the intervention is activated through a reward-gated curriculum. The student is first trained with standard SAC until it acquires basic driving behavior, after which the XAI-guided Fourier perturbation is gradually introduced. The student policy is then optimised under these policy-conditioned and task-consistent visual domains using the SAC objective with auxiliary teacher-guided regularisation terms. The auxiliary teacher-guided terms are used only to stabilise the student response under generated visual domains, while the policy improvement remains driven by the SAC objective.

\subsection{SAC-Based Visual Driving Policy}

The Duckiebot navigation task is modeled as a continuous-control 
Markov decision process (MDP) 
\(\mathcal{M}=(\mathcal{S},\mathcal{A},\mathcal{P},r,\gamma)\). 
At each time step \(t\), the agent receives an observation 
\(o_t=(I_t,n_t)\), where \(I_t\) is the front-view RGB image and 
\(n_t\in\mathbb{R}^{5}\) denotes the low-dimensional navigation state. 
The upper part of the image is cropped before resizing and normalisation 
to reduce task-irrelevant background information. The action 
\(a_t\in\mathbb{R}^{2}\) represents the left and right wheel commands.

We adopt soft actor-critic (SAC)\cite{haarnoja2018softactorcriticoffpolicymaximum} as the base visual control backbone. SAC-style actor-critic formulations have also been used in autonomous-driving decision-making tasks to improve control efficiency and safety under dynamic traffic conditions\cite{9914655}. 
The actor \(\pi_{\theta}(a_t|o_t)\) predicts a stochastic wheel-control 
policy, while two critics \(Q_{\psi_1}\) and \(Q_{\psi_2}\) estimate the 
corresponding soft action values using the clipped double-Q design. 
In our implementation, the actor and critic employ separate visual 
encoders. This separation allows policy and value learning to form 
independent visual representations, and prevents the actor-side 
regularisation introduced later from directly constraining the critic.

This subsection only defines the visual control backbone. The complete 
student optimisation objective under the constructed target-free visual 
domains is given in Section~III-E.

\subsection{Attribution Extraction}

To identify the visual evidence that contributes to the control decision, 
we extract DeepLIFT \cite{shrikumar2019learningimportantfeaturespropagating } attribution maps from a frozen teacher actor. The teacher is instantiated from a domain-randomised SAC policy pretrained in simulation, providing an attribution reference that has already been exposed to visual appearance variations. During XFD training, the teacher network is kept fixed and is used only for attribution extraction and local regularisation.

Given an observation \(o_t=(I_t,n_t)\), the deterministic output of the 
teacher actor is denoted by \(\mu_T(o_t)\in\mathbb{R}^{d_a}\), where 
\(d_a=2\) corresponds to the left and right wheel commands. DeepLIFT 
propagates contribution scores from the teacher action output back to the 
image input relative to a reference activation. For the RGB image \(I_t\), 
the resulting attribution tensor is written as
\begin{equation}
C_T(I_t)\in\mathbb{R}^{H\times W\times 3\times d_a}
\end{equation}

Since the policy output is continuous and multi-dimensional, the 
contributions from both action dimensions are aggregated into a single 
spatial attribution map:
\begin{equation}
\mathcal{A}_T(u,v)=
\frac{1}{Z}
\sum_{c=1}^{3}
\sum_{k=1}^{d_a}
\left|C_T(u,v,c,k)\right|
\end{equation}
where \(Z\) normalizes \(\mathcal{A}_T\) into \([0,1]\). A larger value of 
\(\mathcal{A}_T(u,v)\) indicates that the corresponding image location 
contributes more strongly to the teacher's wheel-control output.

The obtained DeepLIFT attribution map is used as a training-time signal for domain construction, which is consistent with interpretable driving studies that introduce intermediate visual evidence to improve policy transparency\cite{9965619}. Regions with high attribution are treated as 
control-relevant visual evidence, whereas regions with low attribution 
provide larger degrees of freedom for appearance perturbation. This 
attribution-guided separation is passed to the Fourier intervention 
module, allowing the subsequent domain construction to modify visual 
appearance while preserving the image structures most relevant to 
closed-loop control.

\subsection{XAI-Guided Fourier Domain Intervention}

\begin{figure*}[!t]
\centering
\includegraphics[
    width=0.95\textwidth,
    trim=0cm 6.5cm 0cm 4.0cm,
    clip
]{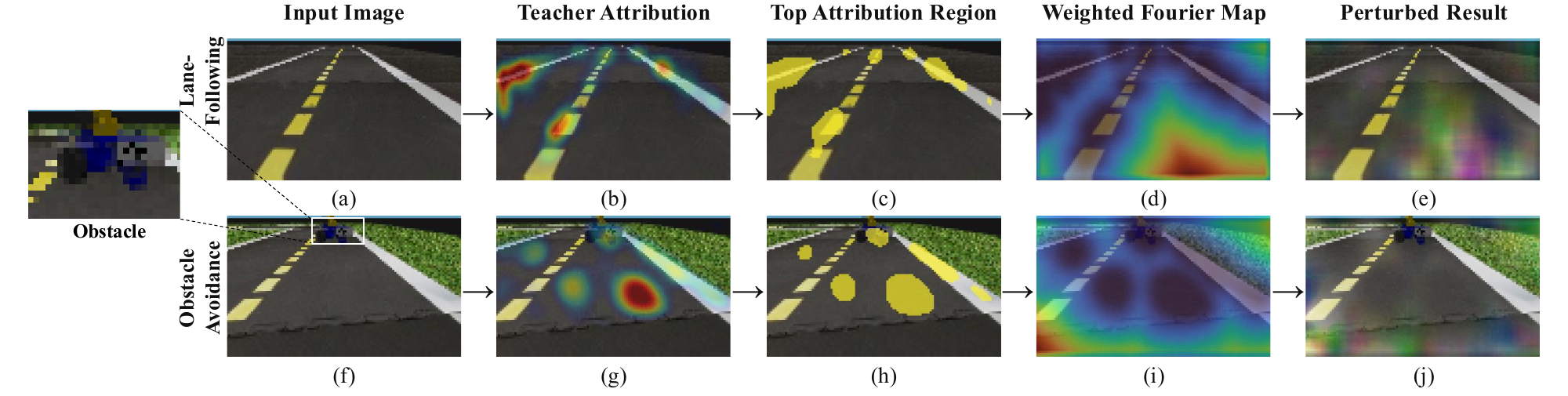}
\caption{Visualisation of the XAI-guided Fourier intervention pipeline under  lane-following and obstacle-avoidance observations. The columns show the input image, teacher attribution, top-attribution region, attribution-weighted Fourier map, and final perturbed result.}
\label{fig:fourier_intervention}
\end{figure*}

The proposed intervention constructs policy-conditioned and task-consistent visual domains by perturbing simulation images in the Fourier domain. The key idea is to alter appearance-dependent statistics while preserving the geometric structures required for closed-loop control. Unlike policy-agnostic domain randomisation~\cite{tobin2017domainrandomizationtransferringdeep} or dynamics randomisation~\cite{Peng_2018}, the perturbation is conditioned on the DeepLIFT attribution map extracted in Section III-C, so that the spatial strength of the intervention is regulated according to control relevance, as illustrated by the teacher attribution responses in Fig.~\ref{fig:fourier_intervention}(b) and Fig.~\ref{fig:fourier_intervention}(g).

Directly applying strong visual perturbations at the beginning of training 
may destabilise exploration, since the student policy has not yet acquired 
a reliable driving prior. Therefore, the intervention is activated by a reward-gated curriculum, following the curriculum-learning principle of introducing stronger interventions after basic task competence has been acquired\cite{8727471}. Let \(R_e\) denote the return of episode \(e\), 
and let \(\bar{R}_e\) be the moving average over the most recent \(K\) 
episodes. The intervention gate is defined as
\begin{equation}
g_t=\mathbb{I}(\bar{R}_e \geq \rho)
\end{equation}
where \(\rho\) is a task-dependent activation threshold. Once the gate is 
enabled at step \(t_g\), the perturbation strength is gradually increased:
\begin{equation}
\eta_t =
g_t \cdot
\min \left(
\eta_{\max},
\eta_{\min}
+
\frac{t-t_g}{T_{\mathrm{ramp}}}
(\eta_{\max}-\eta_{\min})
\right)
\end{equation}
where \(T_{\mathrm{ramp}}\) controls the ramp-up duration, and \(\eta_{\min}\) and \(\eta_{\max}\) denote the minimum and maximum perturbation strengths. This schedule delays the increase of intervention strength until the policy has learned basic control behavior, thereby avoiding abrupt distribution shifts during early optimisation.

Given a preprocessed RGB image $I_t$, as exemplified by the lane-following and obstacle-avoidance inputs in Fig.~\ref{fig:fourier_intervention}(a) and Fig.~\ref{fig:fourier_intervention}(f), the Fourier transform is computed independently for each channel:
\begin{equation}
\mathcal{F}(I_t^c)=A_t^c \exp(jP_t^c),
\quad c\in\{1,2,3\}
\end{equation}
where \(A_t^c\) and \(P_t^c\) denote the amplitude and phase components, 
respectively. The amplitude component is perturbed to modify appearance 
statistics, including illumination, contrast, texture, and color 
distribution, while the phase component is preserved to maintain lane 
geometry, road boundaries, and obstacle layout. The perturbed amplitude is 
defined as
\begin{equation}
\tilde{A}_t^c =
A_t^c \odot
\left(1+\eta_t \Delta_t^c\right)
\end{equation}
where \(\Delta_t^c\) is a bounded spectral perturbation and \(\odot\) 
denotes element-wise multiplication. The intermediate Fourier-perturbed 
image is reconstructed by
\begin{equation}
I_t^{F}=
\mathcal{F}^{-1}
\left(
\tilde{A}_t \exp(jP_t)
\right)
\end{equation}

The top-attribution regions in Fig.~\ref{fig:fourier_intervention}(c) and Fig.~\ref{fig:fourier_intervention}(h) visualise the most action-relevant areas selected from the teacher attribution maps. These binary regions are used only for interpretation, whereas the continuous attribution map is used in the actual intervention. Let $A_T(I_t) \in [0,1]^{H \times W}$ denote the normalised teacher attribution map. The attribution-weighted Fourier shifts shown in Fig.~\ref{fig:fourier_intervention}(d) and Fig.~\ref{fig:fourier_intervention}(i) indicate where the reconstructed appearance variation is allowed to be expressed after being modulated by the teacher attribution. A complementary intervention mask is defined as
\begin{equation}
M_t = 1-\mathcal{A}_T(I_t)
\end{equation}
The final perturbed image is obtained by blending the original image and 
the Fourier-perturbed image:
\begin{equation}
\tilde{I}_t =
I_t+
M_t\odot
\left(
I_t^{F}-I_t
\right)
\end{equation}

This formulation assigns lower perturbation freedom to high-attribution regions and higher perturbation freedom to low-attribution regions. The final perturbed observations in Fig.~\ref{fig:fourier_intervention}(e) and Fig.~\ref{fig:fourier_intervention}(j) show that the generated images change domain-dependent appearance while retaining the lane, boundary, and obstacle structures that are most relevant to control. The intervention therefore converts generic Fourier augmentation into a policy-conditioned visual-domain intervention. The student policy is subsequently trained on $\tilde{o}_t = (\tilde{I}_t, n_t)$, allowing reinforcement learning to optimise control behavior under visually shifted but structurally consistent simulation domains.

\subsection{Target-Free Robust Policy Optimisation}

The student policy is optimised without target-domain images, real-world 
rollouts, or demonstrations. To write the optimisation compactly, we 
define the training observation as
\begin{equation}
x_t =
\begin{cases}
o_t, & g_t=0\\
\tilde{o}_t, & g_t=1
\end{cases}
\end{equation}
where \(g_t\) is the intervention gate defined in Section~III-D. Before 
the intervention is activated, the student is trained on the original 
simulation observation \(o_t\). After activation, it is trained on the 
attribution-guided Fourier observation \(\tilde{o}_t\).

For compact notation, let 
\(Q_{\min}^{\psi}(x,a)=\min_{i=1,2}Q_{\psi_i}(x,a)\) and 
\(Q_{\min}^{\bar{\psi}}(x,a)=\min_{i=1,2}Q_{\bar{\psi}_i}(x,a)\). 
For a transition \((x_t,a_t,r_t,x_{t+1},d_t)\), the SAC target is
\begin{equation}
V_{\bar{\psi}}(x_{t+1}) =
Q_{\min}^{\bar{\psi}}(x_{t+1},a_{t+1})
-\alpha\log\pi_{\theta}(a_{t+1}|x_{t+1})
\end{equation}
where \(a_{t+1}\sim\pi_{\theta}(\cdot|x_{t+1})\). 
The SAC target is then written as
\begin{equation}
y_t = r_t+\gamma(1-d_t)V_{\bar{\psi}}(x_{t+1})
\end{equation}
The critic is optimised by
\begin{equation}
\mathcal{L}_{Q} =
\sum_{i=1}^{2}
\mathbb{E}_{\mathcal{D}}
\left[
\left(Q_{\psi_i}(x_t,a_t)-y_t\right)^2
\right]
\end{equation}
The teacher is not involved in the critic target or critic loss. Thus, 
value learning is driven only by the reinforcement learning objective 
under the constructed domains.

The SAC actor loss is
\begin{equation}
\mathcal{L}_{\mathrm{SAC}} =
\mathbb{E}_{\mathcal{D}}
\left[
\alpha\log\pi_{\theta}(a_t|x_t)
-
Q_{\min}^{\psi}(x_t,a_t)
\right]
\end{equation}
This term allows the student to improve its control behavior through 
reinforcement learning, rather than merely imitating the teacher.

To stabilise policy learning under appearance-shifted observations, a 
behavioral consistency term is introduced between the student response to 
the perturbed observation and the teacher response to the corresponding 
clean observation. Let \(\mu_S(\tilde{o}_t)\) and \(\mu_T(o_t)\) denote 
the deterministic mean actions of the student and teacher actors, 
respectively. The teacher-student consistency loss\cite{10670293} is defined as
\begin{equation}
\mathcal{L}_{\mathrm{cons}} =
\mathbb{E}_{\mathcal{D}}
\left[
\left\|
\mu_S(\tilde{o}_t)
-
\mathrm{sg}\left(\mu_T(o_t)\right)
\right\|_2^2
\right]
\end{equation}
where \(\mathrm{sg}(\cdot)\) denotes the stop-gradient operation. This 
term uses the frozen teacher only as a local behavioral reference and 
does not update the teacher parameters.

In addition to action-level consistency, attribution-level constraints\cite{10475532} 
are applied to regulate the visual evidence used by the student actor. 
Let \(\mathcal{A}_T(I_t)\) and \(\mathcal{A}_S(\tilde{I}_t)\) denote the 
normalised DeepLIFT attribution maps of the teacher and student actors, 
respectively. The outside-attribution loss penalizes student attribution assigned to regions that are weakly supported by the teacher attribution:
\begin{equation}
\mathcal{L}_{\mathrm{out}}
=
\mathbb{E}_{\mathcal{D}}
\left[
\frac{
\left\|
M_t \odot \mathcal{A}_S(\tilde{I}_t)
\right\|_1
}{
\left\|M_t\right\|_1+\epsilon
}
\right]
\end{equation}
where \(M_t=1-\mathcal{A}_T(I_t)\) is the complementary attribution mask 
defined in Section~III-D.
This discourages the student from relying on visually perturbed regions 
that are not strongly supported by the teacher attribution.

A complementary attribution-alignment term preserves the global structure 
of the teacher attribution under the perturbed observation:
Let 
\(\mathbf{a}_T=\mathrm{vec}(\mathcal{A}_T(I_t))\) and 
\(\mathbf{a}_S=\mathrm{vec}(\mathcal{A}_S(\tilde{I}_t))\). 
The attribution-alignment loss is defined as
\begin{equation}
\mathcal{L}_{\mathrm{align}}
=
\mathbb{E}_{\mathcal{D}}
\left[
1-
\frac{
\langle \mathbf{a}_T,\mathbf{a}_S\rangle
}{
\|\mathbf{a}_T\|_2\|\mathbf{a}_S\|_2+\epsilon
}
\right]
\end{equation}

While \(\mathcal{L}_{\mathrm{out}}\) suppresses attribution leakage into 
low-teacher-attribution regions, \(\mathcal{L}_{\mathrm{align}}\) 
encourages the student to preserve the overall spatial attribution 
structure of the teacher.

The final student actor objective combines reinforcement learning, 
behavioral consistency, and attribution regularisation:
\begin{equation}
\begin{aligned}
\mathcal{L}_{\pi} =
\mathcal{L}_{\mathrm{SAC}}
+
g_t \Big(
& \lambda_{\mathrm{cons}}\mathcal{L}_{\mathrm{cons}} \\
& +
\lambda_{\mathrm{xai}}
\left(
\lambda_{\mathrm{out}}\mathcal{L}_{\mathrm{out}}
+
\lambda_{\mathrm{align}}\mathcal{L}_{\mathrm{align}}
\right)
\Big)
\end{aligned}
\end{equation}
where \(g_t\) is the intervention gate defined in Section III-D. When 
\(g_t=0\), the student is optimised by the standard SAC objective. When 
\(g_t=1\), the auxiliary terms regularise the actor toward stable 
action-level and attribution-level behavior under target-free visual 
domain shifts.

\subsection{Obstacle-Aware Semantic Replay}

\begin{figure}[!htbp]
\centering
\includegraphics[
    width=1.05\columnwidth,
    trim= 3.0cm 7cm 2.3cm 0.8cm,
    clip
]{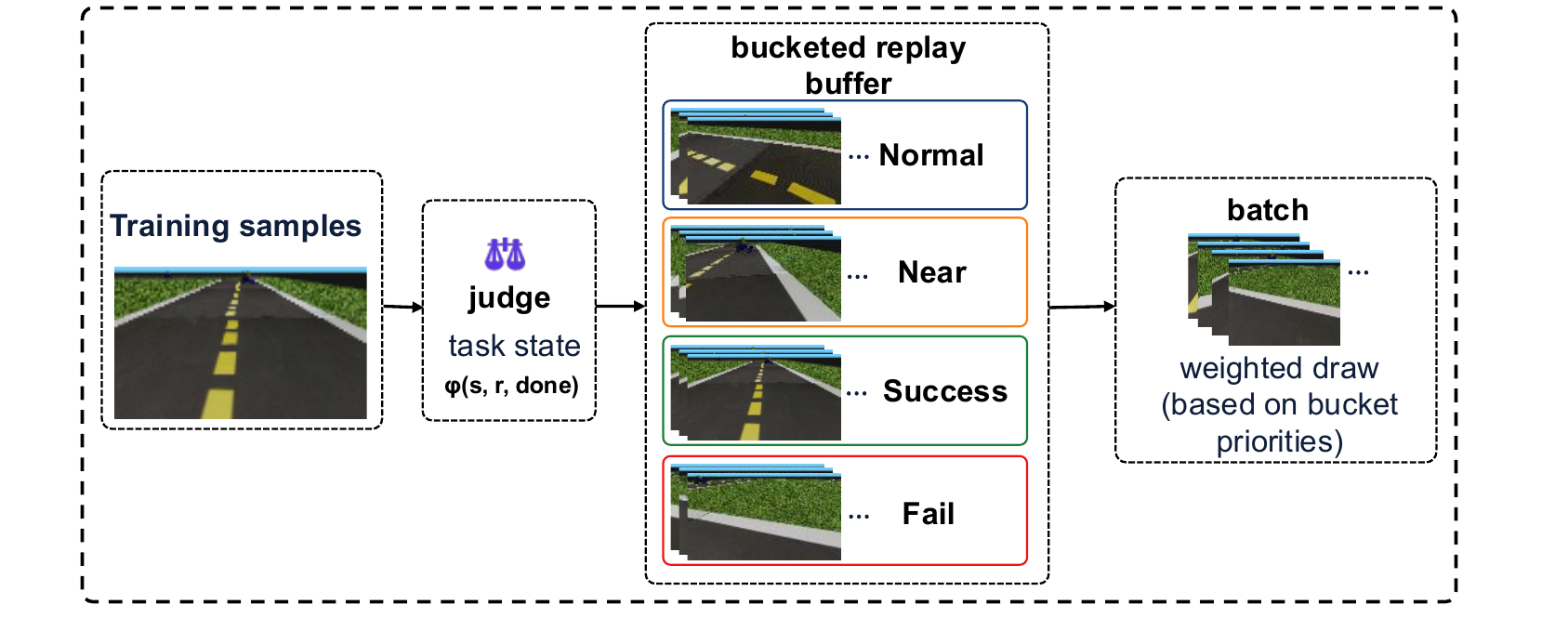}
\caption{Obstacle-aware semantic replay sampling. Transitions collected from randomised obstacle-placement training are assigned to normal, near-obstacle, success, and failure categories, and mini-batches are sampled by weighted category selection.}
\label{fig:semantic_replay}
\end{figure}

Since a single obstacle-aware policy is trained in this work, experience replay is configured for the obstacle-aware training environment rather than for two separately trained task policies. During simulation training, the Duckiebot encounters randomised obstacle placements while still being required to maintain lane-following behavior. Therefore, the replay buffer contains both  lane-tracking transitions and sparse obstacle-interaction transitions. The resulting replay organisation is illustrated in Fig.~\ref{fig:semantic_replay}.

Each collected transition is assigned to one semantic replay category:
\begin{equation}
\mathcal{B}=\{\mathcal{B}_{\mathrm{normal}},\mathcal{B}_{\mathrm{near}},
\mathcal{B}_{\mathrm{success}},\mathcal{B}_{\mathrm{fail}}\},
\end{equation}
where the four categories correspond to normal lane-following transitions, near-obstacle interactions, successful avoidance maneuvers, and failure-related transitions, respectively. The normal category preserves  lane-tracking competence, while the remaining categories increase the exposure of sparse safety-critical obstacle-avoidance cases.

Mini-batches are sampled from a progress-dependent category distribution
\begin{equation}
\omega(p_t)=[\omega_{\mathrm{normal}},\omega_{\mathrm{near}},
\omega_{\mathrm{success}},\omega_{\mathrm{fail}}],
\end{equation}
which satisfies
\begin{equation}
\sum_{b\in\mathcal{B}}\omega_b(p_t)=1,
\end{equation}
where $p_t=t/T$ denotes the normalised training progress. In the early stage, the sampler increases the probability of near-obstacle and failure-related transitions to expose the policy to decision boundaries and unsafe outcomes. As training proceeds, the sampling distribution shifts toward successful avoidance transitions while retaining normal lane-following samples. This design allows the same policy to learn obstacle-aware control without sacrificing  lane-following behavior.

This replay mechanism is independent of the proposed Fourier intervention. 
\begin{figure*}[!t]
\centering
\includegraphics[
    width=0.95\textwidth,
    trim=1cm 6.0cm 1cm 4.8cm,
    clip
]{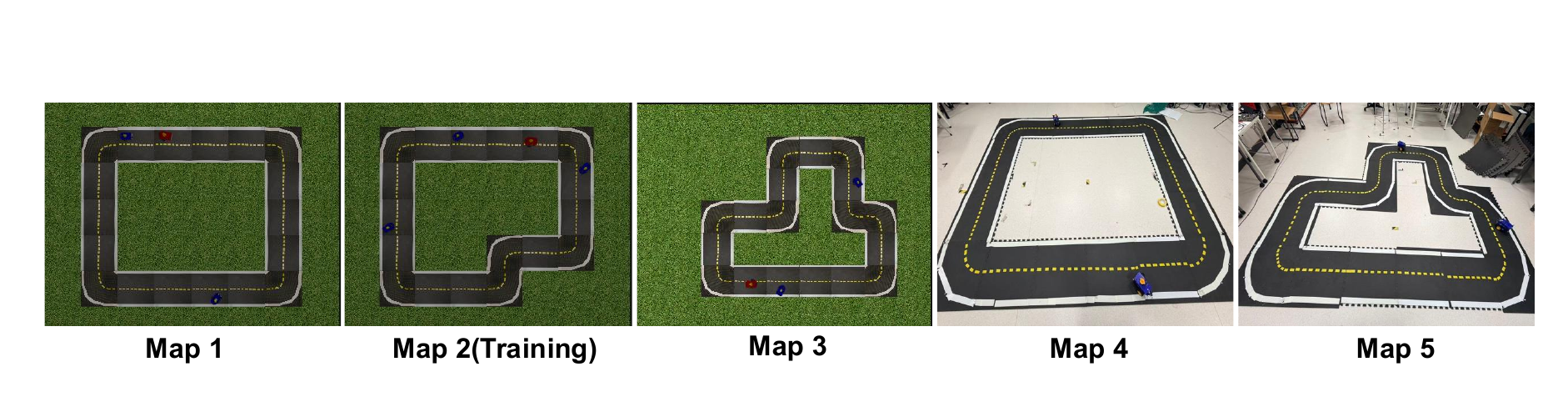}
\caption{Duckietown map layouts for simulation training, simulation evaluation, and zero-shot real-world deployment.}
\label{fig:map_layouts}
\end{figure*}

It acts as a task-level sampling strategy within the same SAC optimisation framework, correcting the long-tailed transition distribution induced by randomised obstacle placement while preserving standard lane-tracking competence.

\subsection{Reward Formulation}

The reward function is used to induce stable lane-following behavior and 
to suppress degenerate policies such as stopping, excessive heading 
deviation, and boundary-hugging. Since reward design is not the main 
contribution of this work, both the baseline and the proposed method are 
trained with the same task reward.

At each timestep, the simulator first checks whether the vehicle can be 
assigned to a valid lane pose. If the lane pose is invalid, a fixed penalty 
\(r_t=-5\) is returned. For valid lane poses, the reward is defined as
\begin{equation}
r_t =
r_t^{\mathrm{prog}}
+
r_t^{\mathrm{move}}
-
p_t^{\mathrm{stop}}
-
p_t^{\mathrm{angle}}
-
p_t^{\mathrm{edge}} 
\end{equation}

The progress reward encourages forward motion along the lane direction:
\begin{equation}
r_t^{\mathrm{prog}}
=
1.5\,s_t\max(c_t,0)
\end{equation}
where \(s_t=\mathrm{clip}(v_t,0,1)\) is the normalised speed and \(c_t\) 
denotes the projection of the vehicle heading onto the lane direction.

A small motion reward is added to discourage stationary behavior:
\begin{equation}
r_t^{\mathrm{move}}
=
0.02\min\left(1,\frac{s_t}{s_{\min}}\right)
\end{equation}
where \(s_{\min}=0.10\).

The stopping penalty is imposed when the normalised speed is lower than 
\(s_{\min}\):
\begin{equation}
p_t^{\mathrm{stop}}
=
\mathbb{I}(s_t<s_{\min})k_{\mathrm{stop}}
\left(\frac{s_{\min}-s_t}{s_{\min}}\right)^2 
\end{equation}

The heading penalty is applied outside an angular dead zone:
\begin{equation}
p_t^{\mathrm{angle}}
=
k_{\mathrm{angle}}\max(0,|\theta_t|-\theta_{\mathrm{dead}})
\end{equation}
where \(\theta_t\) denotes the heading error.

The edge penalty discourages driving close to the lane boundary:
\begin{equation}
p_t^{\mathrm{edge}}
=
k_{\mathrm{edge}}\max(0,|e_t|-e_{\mathrm{thr}})
\end{equation}
where \(e_t\) is the lateral deviation from the lane reference. The 
constants are set to \(k_{\mathrm{stop}}=3.0\), \(k_{\mathrm{angle}}=0.2\), 
\(\theta_{\mathrm{dead}}=30^{\circ}\), \(k_{\mathrm{edge}}=4.0\), and 
\(e_{\mathrm{thr}}=0.10\).

This reward only defines the task objective. The proposed Fourier-domain 
intervention and attribution-guided policy regularisation are independent 
of the reward formulation; therefore, improvements over the baseline are 
attributed to the proposed target-free robust training strategy rather 
than to reward redesign.

\section{Experiments}

This section describes the experimental protocol used to train and evaluate the proposed XAI-guided Fourier domain generalisation framework, including the task and map configurations, baseline methods, training protocols, and evaluation. The evaluation metrics are designed to assess closed-loop robustness, cross-map generalisation, and zero-shot sim-to-real transfer under visual domain shifts. Quantitative comparisons, real-world deployment results, and attribution-based qualitative analyses are presented in Section V.

\subsection{Environment Configurations}
The experiments are conducted in the Duckietown environment using an obstacle-aware training task. During training, the policy is optimised on Map 2, shown in Fig. 4, with randomised obstacle placements. The policy is required to maintain lane-following behaviour while responding appropriately to obstacle interactions.

The experimental map configuration is organised as follows: 
\begin{itemize} \item \textbf{Map 2}: the simulation training map, where the single obstacle-aware policy is trained with randomised obstacle placements. 
\item \textbf{Map 1 and Map 3}: unseen simulation evaluation maps used to assess cross-map generalisation. \item \textbf{Map 4 and Map 5}: physical Duckietown layouts used for zero-shot real-world deployment, corresponding to the evaluation layouts represented by Map 1 and Map 3, respectively. \end{itemize} 
The same trained policies are evaluated on all simulated and real-world maps without target-domain fine-tuning or target domain data acquisition. All the map layouts used in the experiments are demonstrated in Fig.~\ref{fig:map_layouts}.

\subsection{Baseline Methods}

All compared methods use the same SAC-based visual control backbone and are trained under the same obstacle-aware simulation setting on Map 2. The baseline is a standard SAC policy trained without the proposed sim-to-real components. A randomly domain-randomised SAC policy is included to represent the conventional visual randomisation. The proposed XFD policy uses the same backbone and training environment but introduces DeepLIFT-guided Fourier domain intervention and target-free robust policy optimisation during training.

For each method, the same trained policy is evaluated under both lane-following and obstacle-avoidance tasks. This setting keeps the comparison focused on whether the proposed visual-domain intervention improves the robustness of a single obstacle-aware policy, rather than relying on task-specific policy training. For compact notation, the compared methods are denoted as follows: 
\begin{itemize} \item \textbf{Base}: standard SAC without any sim-to-real techniques.
\item \textbf{DR}: conventional domain-randomised SAC with target-free visual randomisation. This model also serves as the teacher policy during XFD training.
\item \textbf{XFD}: the full proposed method. \end{itemize}

\subsection{Training Protocols}

All policies are trained entirely in simulation using Map 2 with randomised obstacle placements. No target-domain images, real-world rollouts, demonstrations, or real-world fine-tuning are used during training. The input observation consists of a cropped front-view RGB image and a five-dimensional navigation state. The action is a two-dimensional continuous wheel command corresponding to the left and right wheel velocities.

The Base and DR policies are trained using the standard SAC framework. For the DR baseline, the visual randomisation parameters are empirically tuned to maximise training performance in simulation. Both policies are trained until the episode reward reaches a stable plateau, indicating policy convergence. XFD is subsequently trained using the converged DR policy as a frozen teacher. The student policy is initially optimised on clean simulation observations using the standard SAC objective. Once the reward-gated criterion is satisfied, the attribution-guided Fourier intervention is activated and the perturbation strength is progressively increased. During this phase, During this phase, the teacher serves only as an attribution source, while only the XFD student policy is updated.

After training, the same policy checkpoint is evaluated without further adaptation on simulation Map 1, Map 2, and Map 3, as well as on the corresponding real-world Map 4 and Map 5. The evaluation includes both  lane-following and obstacle-avoidance tasks. 

\subsection{Evaluation Metrics}

The trained policies are evaluated in both simulation and zero-shot real-world deployment. Each trained checkpoint is evaluated under both lane-following and obstacle-avoidance tasks using the same set of metrics.

For the lane-following task, each method is evaluated over ten runs on each map in the absence of obstacles.  For the obstacle-avoidance task, each map contains two obstacle-placement configurations to increase evaluation diversity. Each configuration is evaluated over five runs, resulting in ten trajectories per map, which are pooled to compute the obstacle-avoidance metrics. The same evaluation protocol is applied in both simulation and real-world experiments. For both tasks, successfully completing one closed-loop circuit of the map is considered a success. A run is terminated if the vehicle leaves the drivable lane or collides with an obstacle.

During real-world evaluation, the Duckiebot pose is recorded using an OptiTrack motion-capture system. The recorded pose sequence is projected onto the calibrated map coordinate frame and resampled at 10 Hz. The resulting map-plane trajectory is used to compute travelled distance, lateral deviation, reconstructed reward, and yaw-based smoothness.

After the trajectories are collected, four complementary metrics are used for quantitative evaluation:
\begin{itemize}
    \item \textbf{Travelled distance}: the cumulative map-plane trajectory length after 10 Hz resampling, reflecting how far the vehicle progresses before termination.
    \item \textbf{Reward}: the cumulative simulator-like return reconstructed from forward speed, heading alignment, and centerline deviation. A terminal failure penalty is added when the vehicle does not complete one lap.
    \item \textbf{Lateral deviation}: the mean absolute distance between the vehicle trajectory and the fitted right-lane centerline, measuring lane-tracking accuracy.
    \item \textbf{Smoothness}: a yaw-based regularity score computed from the measured vehicle yaw acceleration:
\end{itemize}
\begin{equation} S_{\mathrm{smooth}} = \frac{100}{1+\mathrm{RMS}(\ddot{\psi})/r_{\mathrm{ref}}}, \end{equation} 
where $\mathrm{RMS}(\cdot)$ denotes the root mean square operator, $\ddot{\psi}$ denotes the yaw acceleration, and $r_{\mathrm{ref}}$ is a reference normalisation factor. Higher smoothness indicates less abrupt yaw variation. The reported results are presented as the mean and standard deviation across these ten runs.

All methods are evaluated under the same task settings, map layouts, obstacle placements, and reward reconstruction procedure. Therefore, differences in these metrics reflect the effect of the proposed method rather than changes in the evaluation protocol or reward definition. Quantitative results are reported in Section V.

\subsection{Implementation Details}

All policies are implemented in PyTorch using the same SAC-based visual actor-critic architecture. The actor and critic do not share the visual backbone; instead, they use separate MobileNetV2 visual encoders to avoid actor-side regularisation directly constraining critic value estimation. Each RGB observation is encoded by a MobileNetV2-based visual encoder, while the five-dimensional navigation state is processed by a lightweight fully connected navigation encoder. The extracted visual and navigation features are concatenated and passed to the corresponding policy or value head. The actor head outputs the mean and log-standard deviation of a two-dimensional Gaussian wheel-command policy, whereas the critic consists of two independent Q-value heads following the clipped double-Q design. The obstacle-aware student policy is trained on Map~2 with randomised obstacle placements. The main hyper-parameter settings are summarised in Table~\ref{tab:implementation_details}.
\begin{table*}[!t]
\caption{Simulation evaluation results of the obstacle-aware policies under  lane-following and obstacle-avoidance tasks. Results are reported as mean $\pm$ standard deviation over ten evaluation trajectories.}
\label{tab:sim_results}
\scriptsize
\resizebox{\textwidth}{!}{
\begin{tabular}{llccccc}
\toprule
Task& Map & Method & Distance & Reward & Deviation & Smoothness \\
\midrule

\multirow{9}{*}{Lane following}
& \multirow{3}{*}{Map 1}
& Base & $10.95\pm0.01$ & $834.40\pm0.78$ & $\mathbf{0.0333\pm0.0005}$ & $91.44\pm0.32\%$ \\
& & DR (Teacher)  & $10.54\pm0.00$ & $798.46\pm0.36$ & $0.0372\pm0.0003$ & $\mathbf{93.86\pm0.26\%}$ \\
& & XFD (Student) & $\mathbf{11.39\pm0.01}$ & $\mathbf{851.39\pm0.21}$ & $0.0796\pm0.0003$ & $93.23\pm0.12\%$ \\
\cmidrule(lr){2-7}

& \multirow{3}{*}{Map 2}
& Base & $5.88\pm0.02$ & $338.73\pm1.21$ & $\mathbf{0.0460\pm0.0023}$ & $88.92\pm0.41\%$ \\
& & DR (Teacher)  & $10.22\pm0.01$ & $734.70\pm0.77$ & $0.0500\pm0.0003$ & $89.65\pm0.43\%$ \\
& & XFD (Student) & $\mathbf{11.00\pm0.00}$ & $\mathbf{819.31\pm0.36}$ & $0.0766\pm0.0001$ & $\mathbf{90.59\pm0.11\%}$ \\
\cmidrule(lr){2-7}

& \multirow{3}{*}{Map 3}
& Base & $4.04\pm0.02$ & $198.38\pm0.88$ & $\mathbf{0.0504\pm0.0021}$ & $85.88\pm0.51\%$ \\
& & DR (Teacher) & $9.09\pm2.27$ & $587.94\pm195.99$ & $0.0739\pm0.0063$ & $86.53\pm3.40\%$ \\
& & XFD (Student) & $\mathbf{10.72\pm0.01}$ & $\mathbf{799.75\pm0.46}$ & $0.0725\pm0.0003$ & $\mathbf{89.29\pm0.12\%}$ \\
\midrule

\multirow{9}{*}{Obstacle avoidance}
& \multirow{3}{*}{Map 1}
& Base & $11.07\pm0.04$ & $780.54\pm4.19$ & $\mathbf{0.0564\pm0.0005}$ & $85.64\pm0.64\%$ \\
& & DR (Teacher) & $10.42\pm0.01$ & $720.42\pm0.67$ & $0.0677\pm0.0004$ & $\mathbf{90.98\pm0.20\%}$ \\
& & XFD (Student) & $\mathbf{11.36\pm0.01}$ & $\mathbf{802.13\pm0.90}$ & $0.0848\pm0.0019$ & $86.86\pm0.32\%$ \\
\cmidrule(lr){2-7}

& \multirow{3}{*}{Map 2}
& Base & $6.10\pm0.06$ & $277.63\pm7.86$ & $0.0762\pm0.0015$ & $82.62\pm0.61\%$ \\
& & DR (Teacher) & $10.01\pm0.04$ & $689.37\pm7.16$ & $\mathbf{0.0641\pm0.0015}$ & $\mathbf{89.93\pm0.69\%}$ \\
& & XFD (Student) & $\mathbf{10.86\pm0.02}$ & $\mathbf{763.65\pm1.01}$ & $0.0796\pm0.0016$ & $87.96\pm0.18\%$ \\
\cmidrule(lr){2-7}

& \multirow{3}{*}{Map 3}
& Base & $4.13\pm0.01$ & $188.48\pm0.96$ & $\mathbf{0.0733\pm0.0008}$ & $81.89\pm0.50\%$ \\
& & DR (Teacher) & $8.30\pm2.21$ & $447.62\pm201.06$ & $0.1056\pm0.0094$ & $84.68\pm0.81\%$ \\
& & XFD (Student) & $\mathbf{10.96\pm0.01}$ & $\mathbf{759.08\pm3.37}$ & $0.0832\pm0.0004$ & $\mathbf{86.08\pm0.25\%}$ \\
\bottomrule
\end{tabular}
}
\end{table*}
\begin{table}[!t]
\centering
\caption{Hyper-parameter settings used for policy training.}
\label{tab:implementation_details}
\begin{tabular}{ll}
\toprule
\textbf{Item} & \textbf{Setting} \\
\midrule
Training map & Map~2 with randomised obstacles \\
Training steps & $5.0\times10^{5}$ environment steps \\
Replay buffer size & $8.0\times10^{4}$ \\
Batch size & 256 \\
Action repeat & 2 \\
Discount factor $\gamma$ & 0.99 \\
Target smoothing $\tau$ & 0.005 \\
Actor learning rate & $1.0\times10^{-4}$ \\
Critic learning rate & $3.0\times10^{-4}$ \\
Temperature learning rate & $3.0\times10^{-4}$ \\
Fourier activation step & $5.0\times10^{4}$ steps \\
Fourier ramp-up duration & $1.5\times10^{5}$ steps \\
Attribution method & DeepLIFT \\
Deployment format & ONNX \\
Deployment platform & Jetson Nano \\
\bottomrule
\end{tabular}
\end{table}

The attribution-guided Fourier intervention is activated only after the reward-gated condition is satisfied, and the perturbation strength is increased progressively during the ramp-up stage. Attribution regularisation is applied periodically on a small mini-batch during training. During deployment, no Fourier perturbation or DeepLIFT computation is required; only the trained student actor is exported in Open Neural Network Exchange (ONNX) format and deployed on the Duckiebot. Therefore, the proposed method does not introduce additional inference cost during real-world execution.

\section{Results}
\subsection{Simulation Evaluation}

\begin{figure*}[!t]
\centering
\includegraphics[
    width=0.95\textwidth,
    trim=0.4cm 0.4cm 0.4cm 0cm,
    clip
]{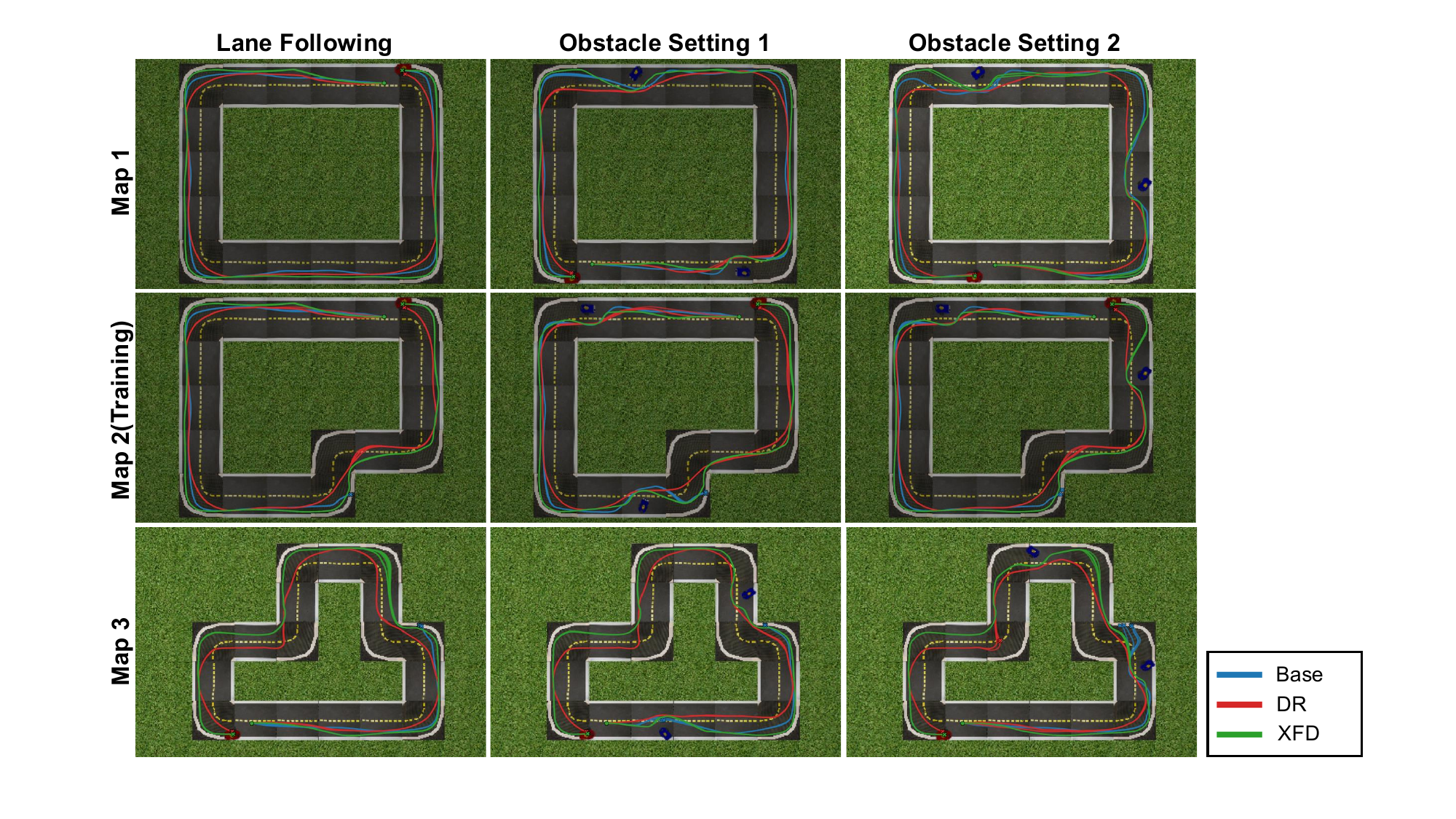}
\caption{Simulation trajectory comparisons on Map 1, Map 2, and Map 3 under lane-following and obstacle-avoidance tasks. Each row corresponds to a simulation map, and each column corresponds to lane following, obstacle setting 1, and obstacle setting 2.}
\label{fig:sim_trajectory}
\end{figure*}

Table~\ref{tab:sim_results} reports the simulation results of the three obstacle-aware policies in lane-following and obstacle-avoidance tasks. All policies are trained only on Map~2 with randomised obstacle placements and are evaluated on Map~1, Map~2, and Map~3 without further adaptation. Map~1 and Map~3 are unseen during training, and therefore mainly reflect cross-map generalisation. The corresponding simulation trajectories are shown in Fig.~\ref{fig:sim_trajectory}.

In the lane-following task, XFD achieves the highest distance and reward across all three maps. On the training map, Map~2, XFD increases the travelled distance from $5.88$ for the base SAC policy and $10.22$ for DR to $11.00$, while improving the reward from $338.73$ and $734.70$ to $819.31$, respectively. The improvement is more pronounced on the unseen Map~3, where the base policy terminates early with a distance of $4.04$ and DR reaches $9.09$ with large variance, whereas XFD maintains a distance of $10.72$ and a reward of $799.75$. This indicates that the proposed visual-domain intervention improves lane-following generalisation even though no separate lane-following policy is trained.

In the obstacle-avoidance task, XFD also obtains the highest distance and reward across all maps. On Map~2, the proposed method improves the distance from $6.10$ for the base policy and $10.01$ for DR to $10.86$. On the unseen Map~3, XFD further increases the distance from $4.13$ and $8.30$ to $10.96$, and improves the reward from $188.48$ and $447.62$ to $759.08$, respectively. These results show that the proposed method improves both lane-tracking behavior and obstacle-aware closed-loop control within the same trained policy.

Averaged over the three simulation maps, XFD improves the lane-following distance by approximately $58.6\%$ over the base policy and $10.9\%$ over DR. For obstacle avoidance, the average distance improvement is approximately $55.8\%$ over the base policy and $15.5\%$ over DR. The corresponding reward gains are also consistent across both tasks, confirming that the proposed method improves task progress rather than only changing low-level trajectory statistics.

These simulation results also highlight the effectiveness of the teacher-student framework in XFD. Although the student receives guidance from the teacher through action and attribution information, it is not trained as an imitation policy. Instead, the teacher provides local stabilisation signals under the generated visual domains, while the student continues to optimise its actor and critic networks through the objective function. Consequently, the student is able to improve beyond the teacher’s performance rather than being constrained by it. This is reflected in Table~\ref{tab:sim_results}: XFD outperforms DR in travelled distance and reward across all simulated maps and both evaluation tasks. Averaged over the three maps, XFD improves the lane-following distance and reward over DR by approximately 10.9\% and 16.5\%, respectively. For obstacle avoidance, the corresponding gains over DR are approximately 15.5\% in distance and 25.2\% in reward. These results show that the proposed method already improves policy performance within simulation, before considering the additional sim-to-real transfer gains.

The deviation and smoothness metrics should therefore be interpreted together with distance and reward. In several cases, the base or DR policy gives lower deviation or higher smoothness, but these values are often obtained from shorter trajectories, conservative motion, or early termination. In contrast, XFD consistently completes longer trajectories and receives higher rewards, which exposes the policy to more curved segments and obstacle-interaction regions. In conclusion, the simulation results indicate that XFD improves closed-loop completion and cross-map robustness while maintaining acceptable trajectory regularity.

\subsection{Real-World Deployment}
\begin{table*}[!t]
\centering
\caption{Zero-shot real-world evaluation results of the single obstacle-aware policy under  lane-following and obstacle-avoidance tasks. Results are reported as mean $\pm$ standard deviation over ten evaluation trajectories.}
\label{tab:real_results}
\scriptsize
\resizebox{\textwidth}{!}{
\begin{tabular}{llccccc}
\toprule
Task& Map & Method & Distance & Reward & Deviation & Smoothness \\
\midrule

\multirow{6}{*}{Lane following}
& \multirow{3}{*}{Map 4}
& Base & $1.37\pm0.31$ & $-92.16\pm15.83$ & $0.2286\pm0.0249$ & $22.54\pm3.83\%$ \\
& & DR (Teacher) & $4.06\pm3.17$ & $133.01\pm210.06$ & $\mathbf{0.0975\pm0.0347}$ & $\mathbf{47.73\pm6.14\%}$ \\
& & XFD (Student)  & $\mathbf{10.26\pm0.12}$ & $\mathbf{576.48\pm16.55}$ & $0.1494\pm0.0054$ & $42.18\pm5.17\%$ \\
\cmidrule(lr){2-7}
& \multirow{3}{*}{Map 5}
& Base & $0.77\pm0.08$ & $-82.53\pm4.69$ & $0.1748\pm0.0193$ & $28.03\pm1.60\%$ \\
& & DR (Teacher)  & $2.91\pm0.87$ & $60.82\pm52.65$ & $\mathbf{0.1270\pm0.0264}$ & $\mathbf{55.52\pm2.42\%}$ \\
& & XFD (Student) & $\mathbf{9.63\pm0.49}$ & $\mathbf{471.17\pm99.60}$ & $0.1537\pm0.0121$ & $42.27\pm4.50\%$ \\
\midrule

\multirow{6}{*}{Obstacle avoidance}
& \multirow{3}{*}{Map 4}
& Base & $1.42\pm0.34$ & $-83.87\pm28.42$ & $0.1318\pm0.0668$ & $24.10\pm7.11\%$ \\
& & DR (Teacher)  & $2.77\pm1.17$ & $61.10\pm81.26$ & $\mathbf{0.0887\pm0.0283}$ & $\mathbf{50.01\pm6.40\%}$ \\
& & XFD (Student) & $\mathbf{9.95\pm0.97}$ & $\mathbf{482.25\pm73.69}$ & $0.1822\pm0.0109$ & $47.32\pm5.12\%$ \\
\cmidrule(lr){2-7}
& \multirow{3}{*}{Map 5}
& Base & $0.98\pm0.40$ & $-92.30\pm37.12$ & $\mathbf{0.1140\pm0.0246}$ & $32.59\pm7.72\%$ \\
& & DR (Teacher)  & $2.85\pm1.39$ & $-3.06\pm65.58$ & $0.1892\pm0.0574$ & $\mathbf{49.35\pm15.71\%}$ \\
& & XFD (Student) & $\mathbf{9.77\pm0.14}$ & $\mathbf{411.28\pm106.57}$ & $0.1873\pm0.0174$ & $44.91\pm2.42\%$ \\
\bottomrule
\end{tabular}
}
\end{table*}

\begin{figure*}[!t]
\centering
\includegraphics[
    width=0.95\textwidth,
    trim=0.2cm 4.5cm 0.2cm 0cm,
    clip
]{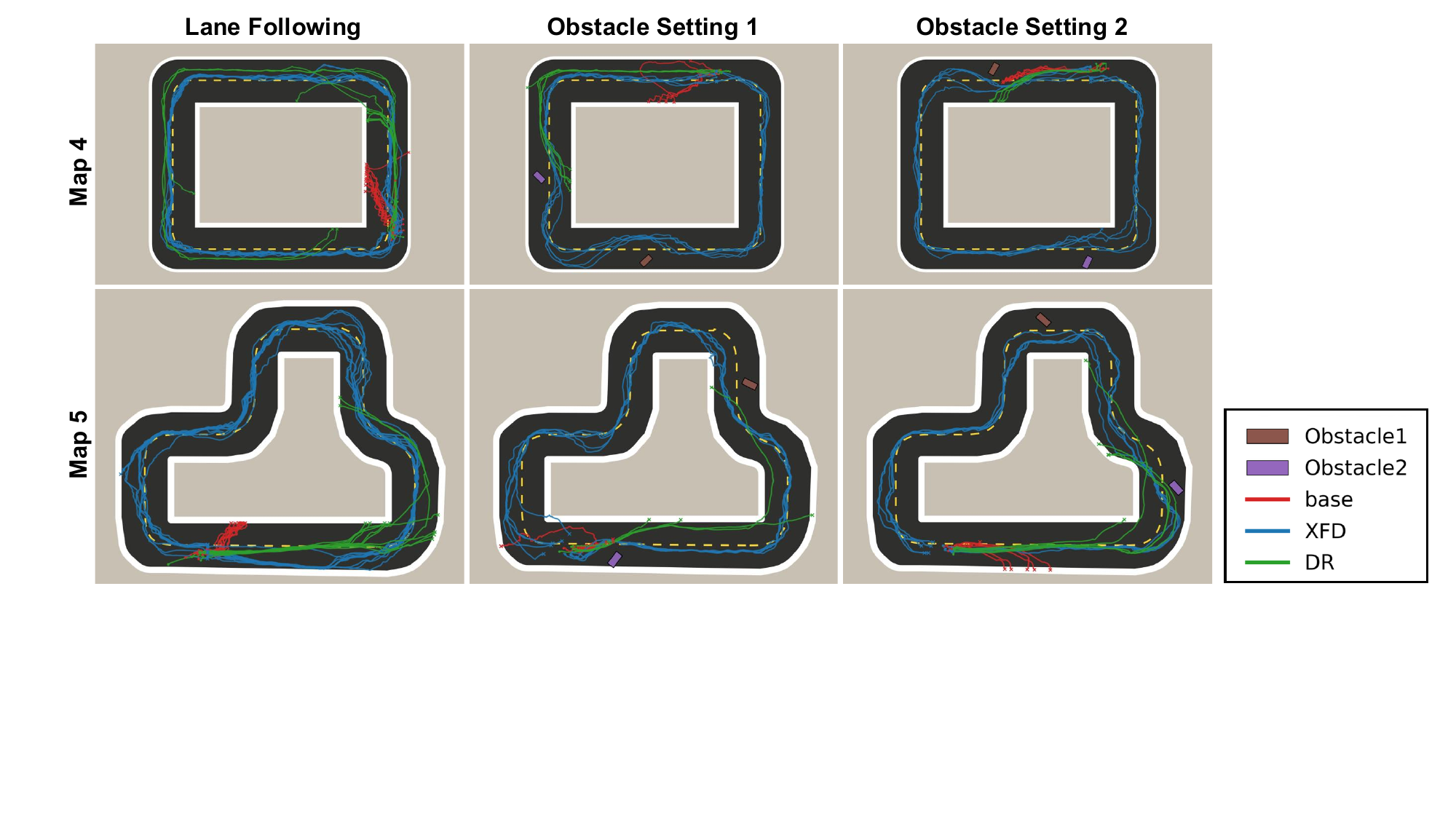}
\caption{Real-world trajectory comparisons on two physical Duckietown maps. The columns correspond to lane following, obstacle setting 1, and obstacle setting 2, respectively.}
\label{fig:real_world_trajectory}
\end{figure*}

Table~\ref{tab:real_results} presents the zero-shot real-world deployment results on physical Map~4 and Map~5. The term target-free refers to the training and adaptation process: the policies are trained entirely in simulation and are deployed without using target-domain images, real-world rollouts, demonstrations, or real-world fine-tuning. The physical maps differ from the simulation training map in layout scale, background appearance, illumination conditions, and robot-environment interaction. Therefore, the real-world evaluation directly tests whether the robustness acquired from the proposed policy-conditioned Fourier intervention transfers to unseen physical visual observations and Duckiebot dynamics.

In the lane-following task, XFD substantially outperforms both the base SAC and DR policies in travelled distance and reward on the two physical maps. Map~4 represents a relatively regular closed-loop layout, where the main challenge comes from real-world appearance changes, camera response, and actuation mismatch. In this setting, XFD reaches a distance of 10.26, compared with 1.37 for the base policy and 4.06 for DR, indicating that the proposed policy can maintain stable lane tracking after direct deployment. Map~5 is more challenging because its compact layout contains consecutive turns and sharper geometric transitions. The base and DR policies terminate early on this map, reaching only 0.77 and 2.91 in distance, respectively, whereas XFD maintains a distance of 9.63. This result shows that the proposed method does not merely adapt to an easy physical layout, but also preserves closed-loop control robustness when the vehicle must continuously adjust its heading through successive turning segments. The reward follows the same trend: the base policy receives negative returns on both maps, DR provides limited improvement, and XFD achieves the highest positive reward. These results demonstrate that the proposed method enables the obstacle-aware policy to maintain robust driving behaviour, both with and without obstacles, despite substantial sim-to-real discrepancies arising from real-world visual shifts, lighting variations, texture discrepancies, background clutter, and differing physical map geometries. This highlights the ability of the proposed framework to learn task-relevant representations that generalise across diverse visual conditions and deployment environments, supporting robust zero-shot transfer from simulation to the physical Duckiebot platform. Furthermore, despite being trained under the guidance of the DR teacher policy, XFD consistently outperforms its teacher across both simulation and real-world evaluations, demonstrating that the proposed framework facilitates policy improvement beyond the capabilities of the teacher rather than merely replicating its behaviour.

\begin{figure*}[!t]
\centering
\includegraphics[
    width=0.95\textwidth,
    trim=1cm 2cm 1.5cm 0.5cm,
    clip
]{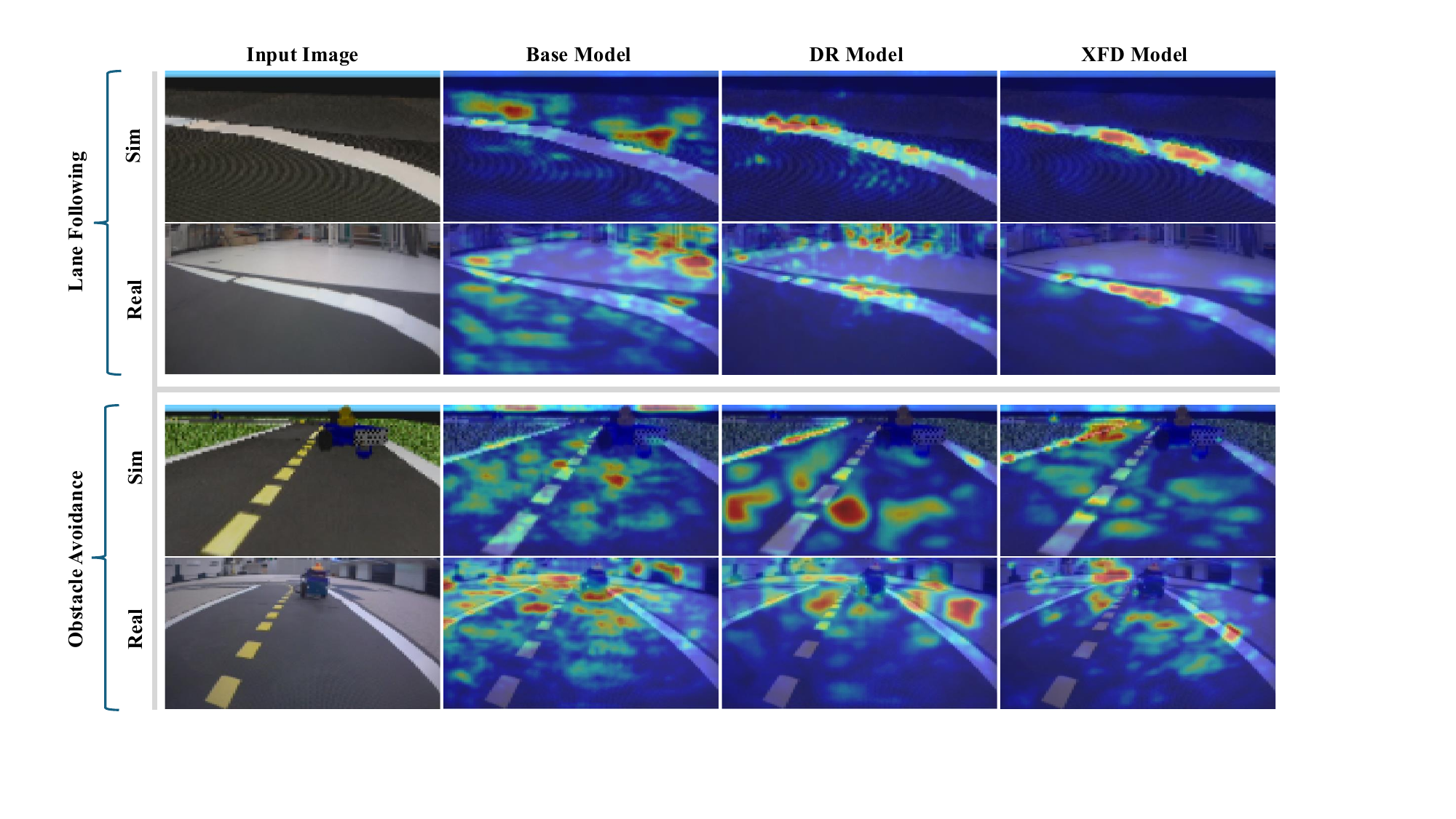}
\caption{Attribution map comparisons among the base model, DR model, and XFD model under two evaluation tasks.}
\label{fig:xai_qualitative}
\end{figure*}
The advantage of XFD is also evident in the obstacle-avoidance task. On Map~4, XFD improves the travelled distance from 1.42 for the base policy and 2.77 for DR to 9.95. On the most difficult Map~5, the distance increases from 0.98 and 2.85 to 9.77. The reward is also markedly improved, rising from negative or near-zero values for the baselines to 482.25 on Map~4 and 411.28 on Map~5. These results demonstrate that XFD enables robust transfer of both lane-following and obstacle-avoidance behaviours to the physical Duckiebot, maintaining reliable performance even in previously unseen environments with consecutive turns and increased navigational complexity. The trajectory visualisation in Fig.~\ref{fig:real_world_trajectory} further reveals a common failure pattern of the base and DR policies. In both physical maps, the baseline policies tend to terminate early around obstacle-interaction or turning regions, where the vehicle must simultaneously maintain lane alignment, avoid the obstacle, and recover to the lane. DR improves visual diversity during training, but its real-world trajectories still show unstable recovery and limited progress after obstacle encounters. In contrast, XFD produces longer and more continuous trajectories across both obstacle settings, suggesting that the proposed policy-conditioned Fourier intervention helps the student preserve task-relevant lane and obstacle cues under real appearance changes.

Averaged over the two real-world maps, XFD improves the lane-following distance from 1.07 for the base policy and 3.49 for DR to 9.95. In obstacle avoidance, the average distance increases from 1.20 and 2.81 to 9.86. Although DR sometimes achieves lower deviation or higher smoothness, its travelled distance and reward remain much lower than those of XFD, indicating that these smoother or more centred trajectories are often associated with conservative motion or early termination rather than successful task completion. Together with Fig.~\ref{fig:real_world_trajectory}, these results demonstrate that the proposed XFD largely improves zero-shot sim-to-real transfer by enabling the trained policy to maintain progress, recover from obstacle interactions, and remain robust across different physical map geometries.

\subsection{XAI Attribution Analysis}

To further investigate the decision-making behavior of the learned visual policies, we visualise the attribution responses of the baseline SAC model, the domain-randomised model, and the XFD model under both simulation and real-world observations. The visualisation is shown in Fig.~\ref{fig:xai_qualitative}. The first column presents the input observations, while the remaining columns show the attribution maps generated by different policies. Brighter regions indicate image areas that contribute more strongly to the corresponding control output.

As shown in Fig.~\ref{fig:xai_qualitative}, the baseline model produces less consistent attribution responses after the transition from simulation to real-world observations. In lane following, its highlighted regions are partially shifted away from the lane structure under real observations, indicating that the policy may rely on visual cues that are sensitive to appearance changes. The domain-randomised model improves visual coverage to some extent, but its attribution is still relatively scattered and includes responses to background or texture-dependent regions. In contrast, the proposed model maintains more concentrated attribution around lane boundaries and road geometry in both simulation and real-world cases. This suggests that the XAI-guided Fourier intervention encourages the student policy to preserve control-relevant visual evidence under visual domain shifts.

A similar tendency can be observed in the obstacle-avoidance task. Compared with lane following, obstacle avoidance requires the policy to jointly process lane geometry, road boundaries, and obstacle-related cues. The baseline and DR models show broader and less stable attribution patterns across simulation and real-world observations, indicating that their action decisions are more easily affected by background texture, appearance changes, or local visual distractors. In contrast, XFD maintains stronger responses around task-relevant regions, including lane markings, road boundaries, and obstacle-related visual cues. This provides a visual explanation for the simulation results in Section~V-A: even before real-world deployment, the student policy improves travelled distance and reward because it learns to rely on more control-relevant evidence rather than merely following the teacher or fitting simulator-specific appearance. 

The same attribution trend also supports the real-world deployment results in Section~V-B. When the observation changes from simulation to physical Duckietown maps, the baseline and DR policies exhibit less consistent evidence selection, which is consistent with their early termination and limited travelled distance in real experiments. By contrast, XFD preserves a more stable attribution structure between simulated and real observations. This suggests that the proposed Fourier intervention improves transfer not only by increasing visual diversity during training, but also by constraining the student policy to keep its action selection grounded in task-critical structures. Therefore, the attribution analysis explains both levels of performance gain: in simulation, XFD improves policy learning beyond the teacher-guided reference; in real-world deployment, it preserves the same task-relevant visual evidence under unseen appearance, lighting, and map-layout changes.

It should be noted that attribution quality is not a direct substitute for closed-loop performance, and a visually plausible attribution map does not by itself guarantee successful control. Nevertheless, prior XAI and explanation-guided learning studies suggest that attribution analysis can reveal whether a policy relies on task-relevant or spurious visual evidence~\cite{10475532,gao2024egl,10716567}. In this work, the quantitative results and attribution maps are consistent: the model with more stable and task-focused visual evidence also achieves longer travelled distance and higher reward in both simulation and real-world deployment. Therefore, the analysis does not claim a universal causal relation between visual explanation and performance, but provides empirical evidence that preserving task-relevant visual evidence is an important factor behind the improved closed-loop robustness of XFD.

\section{Conclusion}
This paper presented XFD, an XAI-guided Fourier domain generalisation framework for target-domain-free sim-to-real reinforcement learning. By leveraging DeepLIFT attributions to regulate Fourier-domain perturbations, the proposed approach generates robustness-enhancing visual domains entirely within simulation, enabling policy learning under structured visual shifts while preserving task-relevant spatial information. To support robust learning, a teacher-student training strategy was introduced, where the teacher provides attribution guidance and training stabilisation while the student remains optimised through the reinforcement learning objective.

Extensive evaluations in simulation and zero-shot real-world deployment demonstrated that XFD consistently outperforms both standard SAC and random domain randomisation baselines across lane-following and obstacle-avoidance tasks. The results demonstrate the complementary benefits of the proposed XAI-guided Fourier intervention and robust policy optimisation strategy. While the Fourier intervention generates policy-relevant visual domains that improve visual generalisation beyond random domain randomisation, the robust optimisation framework enables effective learning under these challenging visual shifts. In combination, XFD improved significantly in closed-loop robustness and sim-to-real transfer across previously unseen environments. Notably, XFD consistently surpasses the DR teacher policy from which it learns in simulation, demonstrating that the teacher serves as a source of guidance and stabilisation rather than a performance upper bound.

Attribution analyses further revealed that XFD promotes more stable and task-focused visual evidence across simulation and real-world observations, supporting the hypothesis that effective sim-to-real transfer should regulate visual shifts according to policy-relevant features rather than relying on indiscriminate visual perturbations. Future work will extend the proposed framework to more complex dynamic environments, larger-scale deployment scenarios, and online attribution-guided adaptation for continual robust policy learning.


\bibliography{sample}
\bibliographystyle{IEEEtran}

\end{document}